\documentclass[twocolumn,a4paper]{article}
\usepackage{nolta2019}
\usepackage{amsmath, bm, enumerate}
\usepackage{txfonts}

\usepackage{my-style}
\usepackage[dvipdfmx]{graphicx}
\usepackage{multirow}
\allowdisplaybreaks[4]

\begin{document}

\title{Improvement of Batch Normalization in Imbalanced Data}

\author{Muneki Yasuda${}^{\S}$ and Seishirou Ueno}

\address{
Graduate School of Science and Engineering, Yamagata University\\
4--3--16 Jyounan, Yonezawa, Yamagata 992-8510, Japan\\
$^{\S}$Email: muneki@yz.yamagata-u.ac.jp
}

\maketitle

\abstract
In this study, we consider classification problems based on neural networks in data-imbalanced environment. 
Learning from an imbalanced data set is one of the most important and practical problems in the field of machine learning. 
A weighted loss function based on cost-sensitive approach is a well-known effective method for imbalanced data sets. 
We consider a combination of weighted loss function and batch normalization (BN) in this study. 
BN is a powerful standard technique in the recent developments in deep learning. 
A simple combination of both methods leads to a size-mismatch problem due to a mismatch between interpretations of effective size of data set in both methods. 
We propose a simple modification to BN to correct the size-mismatch 
and demonstrate that this modified BN is effective in data-imbalanced environment.
\endabstract

\section{Introduction}

Learning from an imbalanced data set is one of the most important and practical problems
and has been actively studied~\cite{ReviewIBD2017,He2009}. 
This study focuses on classification problems based on neural networks in imbalanced data sets. 
Several methods have been proposed for these problems~\cite{He2009} such as under/over-samplings, SMOTE~\cite{SMOTE2002}, 
and cost-sensitive (CS). We consider a CS-based method in this study.
A simple type of CS method is archived by assigning weights to each point of data loss in a loss function~\cite{He2009,CSM2010}. 
Weights in the CS method can be viewed as importances of corresponding data points. 
In an imbalanced data set, the weights of data points that belonging to majority classes are set to smaller than those that belonging to minority classes. 
As discussed in Sec.~\ref{sec:WLF}, the weighting changes the interpretation of the effective size of data set in the resultant weighted loss function.

Batch normalization (BN) is a powerful regularization method for neural networks~\cite{BN2015} 
and it has been one of the most important techniques in deep learning~\cite{DL2016}. 
In the BN of a specific mini-batch, affine signals from units in lower layers are normalized over the data points in the mini-batch. 
A simple combination of the weighted loss function and BN causes a size-mismatch problem.  
As mentioned above, the interpretation of the effective size of data set is changed in the weighted loss function.  
Therefore, the effective size of the mini-batch should be changed in accordance with it. 
In this study, we propose a simple modification to BN to correct the size-mismatch. 

The remainder of this paper is organized as follows. 
In Sec.~\ref{sec:WLF} and \ref{sec:BN}, we briefly explain the weighted loss function and BN, respectively. 
In Sec.~\ref{sec:proposed}, we demonstrate the proposed modification on BN 
and show the results of experiments using MNIST which is a database comprising labeled images of handwritten digits. 
The results show that our method is effective in data-imbalanced environment.
Finally, the conclusion is presented in Sec.~\ref{sec:conclusion}.

\section{Weighted Loss Function}
\label{sec:WLF}

Let us consider a classification model that classifies a $n$-dimensional input vector $\bm{x} = (x_1, x_2,\ldots, x_n)^{\mrm{T}}$ into $K$ different classes 
$C_1, C_2, \ldots, C_K$. 
It is convenient to use 1-of-$K$ representation (or 1-of-$K$ coding) to identify each class~\cite{Bishop2006}.
In 1-of-$K$ representation, each class corresponds to the $K$ dimensional vector $\bm{t} = (t_1, t_2,\ldots, t_K)^{\mrm{T}}$ whose elements are 
$t_k \in \{0,1\}$ and $\sum_{k = 1}^K t_k = 1$, i.e., a vector in which only one element is one and the remaining elements are zero. 
When $t_k = 1$, $\bm{t}$ indicates class $C_k$. 
For the simplicity of notation, we denote the 1-of-$K$ vector, whose $k$th element is one, by $\bm{1}_k$, 
so that $\bm{t} \in T_K := \{\bm{1}_k \mid k = 1,2,\ldots,K\}$.

Suppose that a training data set comprising $N$ data points: 
$\mcal{D}:= \{(\mbf{x}^{(\mu)}, \mbf{t}^{(\mu)}) \mid \mu \in \Omega :=\{ 1,2,\ldots,N\}\}$, 
where $\mbf{x}^{(\mu)} \in \mathbb{R}^n$ and $\mbf{t}^{(\mu)} \in T_K$ are $\mu$th input and corresponding (1-of-$K$ represented) target-class label, respectively, is given.   
Let us consider the minimizing problem of the loss function given by:  
\begin{align}
L(\theta):= \frac{1}{N}\sum_{\mu \in \Omega} f(\mbf{x}^{(\mu)}, \mbf{t}^{(\mu)}; \theta),
\label{eqn:LossFunction}
\end{align}
where $f(\mbf{x}^{(\mu)}, \mbf{t}^{(\mu)}; \theta)$ expresses the loss (e.g., the cross-entropy loss) for $\mu$th data point 
and $\theta$ denotes the set of learning parameters in the classification model.

A weighted loss function is introduced as:
\begin{align}
L_{\mrm{w}}(\theta):= \frac{1}{Z}\sum_{ \mu \in \Omega} w_{\mu} f(\mbf{x}^{(\mu)}, \mbf{t}^{(\mu)}; \theta),
\label{eqn:WeightedLossFunction}
\end{align}
where $w_{\mu} >  0$ is the weight of $\mu$th data point and $Z := \sum_{\mu \in \Omega}w_{\mu}$.
It can be viewed as a simple type of CS approach~\cite{He2009,CSM2010}. 
For an imbalanced data set, the weights are often set to the inverse class frequency~\cite{CSM2016,CSM2017,CSM2018}. 
Here, we assume that $\mcal{D}$ is an imbalanced data set, implying that the sizes of $N_k$s are imbalanced 
(some $N_k$s are very large (\textit{majority} classes) and some are very small (\textit{minority} classes)), 
where $N_k$ is the number of data points belonging to class $C_k$, i.e., $N_k := \sum_{\mu \in \Omega}\delta(\bm{1}_k, \mbf{t}^{(\mu)})$, 
where $\delta(\bm{a},\bm{b})$ is the Kronecker delta function.  
In this case, the weights are set to
\begin{align}
w_{\mu}^{-1} = \frac{1}{N}\sum_{k=1}^K N_k \delta(\bm{1}_k,\mbf{t}^{(\mu)}).
\label{eqn:InverseWeight}
\end{align}
The weights in Eq.~(\ref{eqn:InverseWeight}) effectively correct the imbalance in the loss function due to the imbalanced data set. 
In the weighted loss function with Eq.~(\ref{eqn:InverseWeight}), $\mu$th data point that belongs to class $\mcal{C}_k$ is replicated to $w_{\mu} = N / N_{k}$ data points, 
in which the effective sizes of $N_k$ becomes
\begin{align}
N_k^{\mrm{effective}} := \sum_{\mu \in \Omega} w_{\mu}\delta(\bm{1}_k, \mbf{t}^{(\mu)}) = N.
\end{align}
This means that the effective sizes of $N_k$ (i.e., $N_k^{\mrm{effective}}$) are balanced in the weighted loss function.

\section{Batch Normalization}
\label{sec:BN}

Consider a standard neural network whose $\ell$th layer consists of $h_{\ell}$ units.
In the standard scenario of the feed-forward propagation for input $\mbf{x}^{(\mu)}$ in the neural network, 
the $j$th unit in the $\ell$th layer receives an affine signal from the units in the $(\ell-1)$th layer as: 
\begin{align}
u_{j,\mu}^{(\ell)} := \sum_{i=1}^{h_{\ell-1}}W_{j,i}^{(\ell)}z_{i,\mu}^{(\ell-1)},
\label{eqn:Affine_NN}
\end{align}
where $W_{j,i}^{(\ell)}$ is the connection-weight parameter from the $i$th unit in the $(\ell-1)$th layer to the $j$th unit in the $\ell$th layer 
and $z_{i,\mu}^{(\ell-1)}$ is the output signal of the $i$th unit in the $(\ell-1)$th layer. 
After receiving the signal, the $j$th unit in the $\ell$th layer outputs: 
\begin{align}
z_{j,\mu}^{(\ell)} := a_{\ell}\left(b_j^{(\ell)} + u_{j,\mu}^{(\ell)}\right),
\label{eqn:Output_NN}
\end{align}
where $b_j^{(\ell)}$ is the bias parameter of the unit and $a_{\ell}(x)$ is the specific activation function of the $\ell$th layer.

For BN, the training data is divided into $R$ mini-batches: $B_1,B_2,\ldots,B_R \subseteq \Omega$. 
BN is based on mini-batch-wise stochastic gradient descent (SGD). 
During training with BN, in the feed-forward propagation for input $\mbf{x}^{(\mu)}$ in mini-batch $B_r$, 
Eqs.~(\ref{eqn:Affine_NN}) and (\ref{eqn:Output_NN}) are replaced by:
\begin{align}
\hat{u}_{j,\mu}^{(\ell)} := \gamma_j^{(\ell)}\left(\frac{\lambda_{j,\mu}^{(\ell)} - m_{j,r}^{(\ell)}}{\sqrt{v_{j,r}^{(\ell)} +\varepsilon }}\right),
\quad \lambda_{j,\mu}^{(\ell)} = \sum_{i=1}^{h_{\ell-1}}W_{j,i}^{(\ell)}\hat{z}_{i,\mu}^{(\ell-1)}
\label{eqn:Affine_NN_BN}
\end{align}
and 
\begin{align}
\hat{z}_{j,\mu}^{(\ell)} := a_{\ell}\left(b_j^{(\ell)} + \hat{u}_{j,\mu}^{(\ell)}\right),
\label{eqn:Output_NN_BN}
\end{align}
respectively, where 
\begin{align}
m_{j,r}^{(\ell)}:=\frac{1}{|B_r|}\sum_{\mu \in B_r}\lambda_{j,\mu}^{(\ell)}
\label{eqn:affine_expectation}
\end{align}
and
\begin{align}
v_{j,r}^{(\ell)}:=\frac{1}{|B_r|-1}\sum_{\mu \in B_r}\left(\lambda_{j,\mu}^{(\ell)} - m_{j,r}^{(\ell)}\right)^2
\label{eqn:affine_variance}
\end{align}
are the expectation and (unbiased) variance of the affine signals over the training inputs in $B_r$, and $|B_r|$ denotes the size of $B_r$. 
It should be noted that $\varepsilon$ in Eq.~(\ref{eqn:Affine_NN_BN}) is a small positive value to avoid the division by zero and it is usually set to $10^{-8}$. 
At each layer, the distribution of affine signals, $\{\lambda_{j,\mu}^{(\ell)}\}$, is standardized by Eq.~(\ref{eqn:Affine_NN_BN}). 
It is noteworthy that in training with BN, $\{\gamma_j^{(\ell)}\}$ in Eq.~(\ref{eqn:Affine_NN_BN}) are also the learning parameters with $\{b_j^{(\ell)}\}$ and $\{W_{j,i}^{(\ell)}\}$, 
which are determined by an appropriate back-propagation.
In the inference stage for a new input, we use
$m_{j}^{(\ell)}:=R^{-1}\sum_{r=1}^R m_{j,r}^{(\ell)}$ and $v_{j}^{(\ell)}:=R^{-1}\sum_{r=1}^R v_{j,r}^{(\ell)}$ (i.e., the average values over mini-batches) 
instead of Eqs.~(\ref{eqn:affine_expectation}) and (\ref{eqn:affine_variance}), respectively.

\section{Proposed Method and Numerical Experiment}
\label{sec:proposed}

In this section, we consider BN when we employ the weighted loss function in Eq.~(\ref{eqn:WeightedLossFunction}).
The idea of our method is simple.
In BN, the affine signals are normalized on the basis of sizes of mini-batches.
As mentioned in Sec.~\ref{sec:WLF}, in the weighted loss function, the interpretation of the effective size of data set is changed in accordance with the corresponding weights. 
Thus, to maintain a consistency with the reinterpretation in BN, the sizes of mini-batches should be reviewed. 
This implies that 
\begin{align}
\tilde{m}_{j,r}^{(\ell)}:=\frac{1}{Z_{r}}\sum_{\mu \in B_r} w_{\mu} \lambda_{j,\mu}^{(\ell)}
\label{eqn:affine_expectation_new}
\end{align}
and
\begin{align}
\tilde{v}_{j,r}^{(\ell)}:=\frac{1}{Z_{r}-1}\sum_{\mu \in B_r}w_{\mu}\left(\lambda_{j,\mu}^{(\ell)} - \tilde{m}_{j,r}^{(\ell)}\right)^2
\label{eqn:affine_variance_new}
\end{align}
should be used in Eq.~(\ref{eqn:Affine_NN_BN}) instead of Eqs.~(\ref{eqn:affine_expectation}) and (\ref{eqn:affine_variance}), 
where
\begin{align*}
Z_{r}:=\sum_{\mu \in B_r}w_{\mu}.
\end{align*}
Because our proposed method is realized by replacing Eqs.~(\ref{eqn:affine_expectation}) and (\ref{eqn:affine_variance}) by 
Eqs.~(\ref{eqn:affine_expectation_new}) and (\ref{eqn:affine_variance_new}), respectively, 
it does not produce any additive hyper-parameters. 
This replacement slightly changes the form of back-propagation.  
However, these details are omitted owing to space limitations.

Next, the results of our numerical experiments are illustrated. 
In the experiments, we used the data set available in MNIST, which is a database of handwritten digits, $0, 1,\ldots,9$. 
Each data point includes the input data, a $28 \times 28$ digit image, and the corresponding target digit label. 
Using MNIST, we executed three types of two-class classification problems with \textit{strongly imbalanced} data: 
(i) the classification problem of ``zero'' and ``one'', (ii) the classification problem of ``one'' and ``eight'', 
and (iii) the classification problem of ``four'' and ``seven''. 
The number of training data and test data points used in these experiments are shown in Tab.~\ref{tab:number_of_data}, which were randomly picked up from MNIST. 
For the experiments, we used a three-layered neural network with 784 input units, 200 hidden units with rectified linear function (ReLU) activation~\cite{ReLU2011}, 
and two soft-max output units. We adopted BN for the hidden and output layers.
In the training of neural network, we used the Xavier's initialization~\cite{Xavier2010}, the Adam optimizer~\cite{Adam2015}, 
and set $|B_r| = 100$. 
\begin{table}[]
\centering
\caption{Number of data points used in experiments (i), (ii), and (iii). 
In each experiment, the minority data accounts for less than 1\% of all data.}
\begin{tabular}{cc|c|c|}
\cline{3-4}
\multicolumn{1}{l}{} & \multicolumn{1}{l|}{} & \multicolumn{1}{l|}{train} & \multicolumn{1}{l|}{test} \\ \hline
\multicolumn{1}{|c|}{\multirow{2}{*}{(i)}} & 0 (majority)& 5923 & 980 \\ \cline{2-4} 
\multicolumn{1}{|c|}{} & 1 (minority)& 45 & 1135 \\ \hline
\multicolumn{1}{|c|}{\multirow{2}{*}{(ii)}} & 1 (minority)& 45 & 1135 \\ \cline{2-4} 
\multicolumn{1}{|c|}{} & 8 (majority)& 5851 & 974 \\ \hline
\multicolumn{1}{|c|}{\multirow{2}{*}{(iii)}} & 4 (minority)& 39 & 982 \\ \cline{2-4} 
\multicolumn{1}{|c|}{} & 7 (majority)& 6265 & 1028 \\ \hline
\end{tabular}
\label{tab:number_of_data}
\end{table}

The results of the three experiments, (i), (ii), and (iii), are shown in Tab.~\ref{tab:classification_rate}. 
For each experiments, we used three different methods: 
(a) employing the standard loss function in Eq.~(\ref{eqn:LossFunction}) and using the standard BN explained in Sec.~\ref{sec:BN} (LF+sBN), 
(b) employing the weighted loss function in Eq.~(\ref{eqn:WeightedLossFunction}) and using the standard BN (WLF+sBN),    
and (c) employing the weighted loss function in Eq.~(\ref{eqn:WeightedLossFunction}) and using the proposed BN (WLF+pBN). 
The weights in the weighted loss function were set according to Eq.~(\ref{eqn:InverseWeight}). 
The proposed method (WLF+pBN) was found to be the best in all experiments.
\begin{table*}[]
\centering
\caption{Classification accuracies (of each digit and overall) for the test set.}
\begin{tabular}{c|r|r|r|r|r|r|r|r|r|}
\cline{2-10}
\multicolumn{1}{l|}{} & \multicolumn{3}{c|}{(i)} & \multicolumn{3}{c|}{(ii)} & \multicolumn{3}{c|}{(iii)} \\ \cline{2-10} 
\multicolumn{1}{l|}{} & \multicolumn{1}{c|}{0} & \multicolumn{1}{c|}{1} & \multicolumn{1}{c|}{overall} & \multicolumn{1}{c|}{1} & \multicolumn{1}{c|}{8} & \multicolumn{1}{c|}{overall} & \multicolumn{1}{c|}{4} & \multicolumn{1}{c|}{7} & \multicolumn{1}{c|}{overall} \\ \hline
\multicolumn{1}{|c|}{(a) LF+sBN} & 100\% & 67.4\% & 82.5\% & 71\% & 100\% & 84.4\% & 34.8\% & 100\% & 68.2\% \\ \hline
\multicolumn{1}{|c|}{(b) WLF+sBN} & 100\% & 63.3\% & 80.28\% & 75\% & 100\% & 87.9\% & 28.1\% & 100\% & 64.9\% \\ \hline
\multicolumn{1}{|c|}{(c) WLF+pBN (proposed)} & 99.8\% & 87.5\% & \textbf{93.2}\% & 83\% & 100\% & \textbf{92.7\%} & 49.3\% & 99.4\% & \textbf{74.9\%} \\ \hline
\end{tabular}
\label{tab:classification_rate}
\end{table*}

\section{Concluding Remarks}
\label{sec:conclusion}

In this paper, we proposed a modification in BN for the weighted loss function 
and applied our method to imbalanced data sets. 
The idea of the proposed method is simple but is essential. 
Our method improved the classification accuracies in the experiments.

Recently, a new type of weighted loss function (referred to as class-balanced loss function), which is effective for imbalanced data, was proposed~\cite{CBL2019}. 
In this function, the weights are set to:
\begin{align}
w_{\mu}=\frac{1-\beta}{1-\beta^{\alpha_{\mu}}},
\label{eqn:CBLoss}
\end{align}
for $\beta \in [0,1)$, where $\alpha_{\mu}:=\sum_{k=1}^K N_k \delta(\bm{1}_k,\mbf{t}^{(\mu)})$. The class-balanced loss function is equivalent to Eq.~(\ref{eqn:LossFunction}) when $\beta = 0$, 
and the weighted loss function with inverse-class-frequency weights used in this study when $\beta \to 1$, except for the difference in constant factor. 
The forms of $w_{\mu}$s are not limited in our method, which is then available in the class-balanced loss function. 
We will address this in future studies.

\section*{Acknowledgments}
This work was partially supported by JSPS KAKENHI (Grant Numbers: 15H03699, 18K11459, and 18H03303), 
JST CREST (Grant Number: JPMJCR1402), and the COI Program from the JST (Grant Number JPMJCE1312).  

\bibliographystyle{unsrt}
\bibliography{citation}

\end{document}